\documentclass[conference]{IEEEtran}
\IEEEoverridecommandlockouts
\usepackage{cite}
\usepackage{amsmath,amssymb,amsfonts}
\usepackage{algorithmic}
\usepackage{graphicx}
\usepackage{textcomp}
\usepackage{xcolor}
\def\BibTeX{{\rm B\kern-.05em{\sc i\kern-.025em b}\kern-.08em
    T\kern-.1667em\lower.7ex\hbox{E}\kern-.125emX}}
\begin{document}

% \title{BrainDecoder: Style-Based Visual Decoding\\ of EEG Signals}
\title{BrainDecoder: Style-Based Visual Decoding\\ of EEG Signals\thanks{This work has been submitted to the IEEE for possible publication. Copyright may be transferred without notice, after which this version may no longer be accessible.}}

\author{\IEEEauthorblockN{Minsuk Choi}
    \IEEEauthorblockA{\textit{Department of Computer Science and Engineering} \\
        \textit{Waseda University}\\
        Tokyo, Japan \\
        minsuk@fuji.waseda.jp}
    \and
    \IEEEauthorblockN{Hiroshi Ishikawa\thanks{This work was partially supported by JSPS KAKENHI Grant Number JP20H00615.}}
    \IEEEauthorblockA{\textit{Department of Computer Science and Engineering} \\
        \textit{Waseda University}\\
        Tokyo, Japan \\
        hfs@waseda.jp}
}
\maketitle

\begin{abstract}
    Decoding neural representations of visual stimuli from electroencephalography (EEG) offers valuable insights into brain activity and cognition. Recent advancements in deep learning have significantly enhanced the field of visual decoding of EEG, primarily focusing on reconstructing the semantic content of visual stimuli. In this paper, we present a novel visual decoding pipeline that, in addition to recovering the content, emphasizes the reconstruction of the style, such as color and texture, of images viewed by the subject. Unlike previous methods, this ``style-based'' approach learns in the CLIP spaces of image and text separately, facilitating a more nuanced extraction of information from EEG signals. We also use captions for text alignment simpler than previously employed, which we find work better. Both quantitative and qualitative evaluations show that our method better preserves the style of visual stimuli and extracts more fine-grained semantic information from neural signals. Notably, it achieves significant improvements in quantitative results and sets a new state-of-the-art on the popular Brain2Image dataset.
\end{abstract}


% \vspace{1pt}

\section{Introduction}

%The human brain is an intricate and powerful organ capable of efficiently %processing vast amounts of information from the surrounding environment. One of the %most remarkable aspects of brain function is the visual system, which allows us to %interpret the world through sight. This system involves complex neural networks %that transform visual input from light waves to low level features to meaningful %perceptions.

Understanding neural representations in the brain and the information they encode is crucial for enhancing our knowledge of cognitive processes and developing brain-computer interfaces (BCIs) \cite{bcisurvey}. In particular, decoding and simulating the human visual system has emerged as a significant challenge. Recent advancements have led to substantial progress in visual decoding, allowing for the reconstruction of visual stimuli perceived by a subject during brain activity measurement. \cite{brain2image} \cite{thoughtviz} \cite{neurovision} \cite{fang2020reconstructing} \cite{eeg2image} \cite{EEGStyleGAN-ADA} \cite{mindvis} \cite{takagi2023high}

Electroencephalography (EEG) is a technique for recording brain signals, widely used due to its non-invasive nature, cost-effectiveness, and high temporal resolution. 
Although it has notable limitations \cite{eegartifact} such as relatively lower spatial resolution as well as susceptibility to physiological artifacts and individual differences,
%However, EEG has notable limitations \cite{eegartifact}. For instance, compared to functional magnetic resonance imaging (fMRI), EEG signals have a much lower spatial resolution, making it challenging to pinpoint the exact sources of brain activity. Furthermore, EEG signals are often affected by physiological artifacts and individual differences, which complicates the extraction of relevant neural information and hinders accurate decoding. Despite these challenges
conducting research based on EEG remains crucial for practical applications. The technique's accessibility and ability to capture real-time brain activity make it invaluable.
% for developing brain-computer interfaces, monitoring neurological conditions, and advancing our understanding of cognitive processes in real-world settings.

Previous research \cite{brainvis} \cite{dreamdiffusion} \cite{neuroimagen} \cite{li2024visual} on EEG-based visual decoding has primarily focused on capturing high-level semantic content by aligning with the text or image embedding space of CLIP (Contrastive Language–Image Pretraining) \cite{clip}. While these approaches have successfully represented broad semantic categories, they often fall short in accurately reproducing stylistic details such as color and texture, revealing a gap between semantic understanding and detailed visual representation.

In this paper, we present \textit{\textbf{BrainDecoder}}, a novel method that aims to overcome this limitation by aligning EEG signals with both image and text embeddings as separate conditions in a pretrained latent diffusion model \cite{stablediffusion}. In the text-to-image generation literature, previous researches \cite{ipadapter} \cite{instantid} have demonstrated that incorporating image ``prompts'' along with the text ones enable image generation that preserves style and content. By aligning EEG signals with both image and text embedding spaces, we show it is possible to extract both style and semantic information. This dual approach enhances the model's ability to more accurately reconstruct the stylistic features of the images viewed by the EEG subject. Our qualitative and quantitative evaluations demonstrate that BrainDecoder outperforms the state-of-the-art by a large margin in both reconstruction details and generation quality, setting a new benchmark for EEG-based visual decoding.

\begin{figure*}[htbp]
    % \centerline{\includegraphics[width=\textwidth]{Fig_Framework.png}} % for icassp
    \centerline{\includegraphics[width=0.95\textwidth]{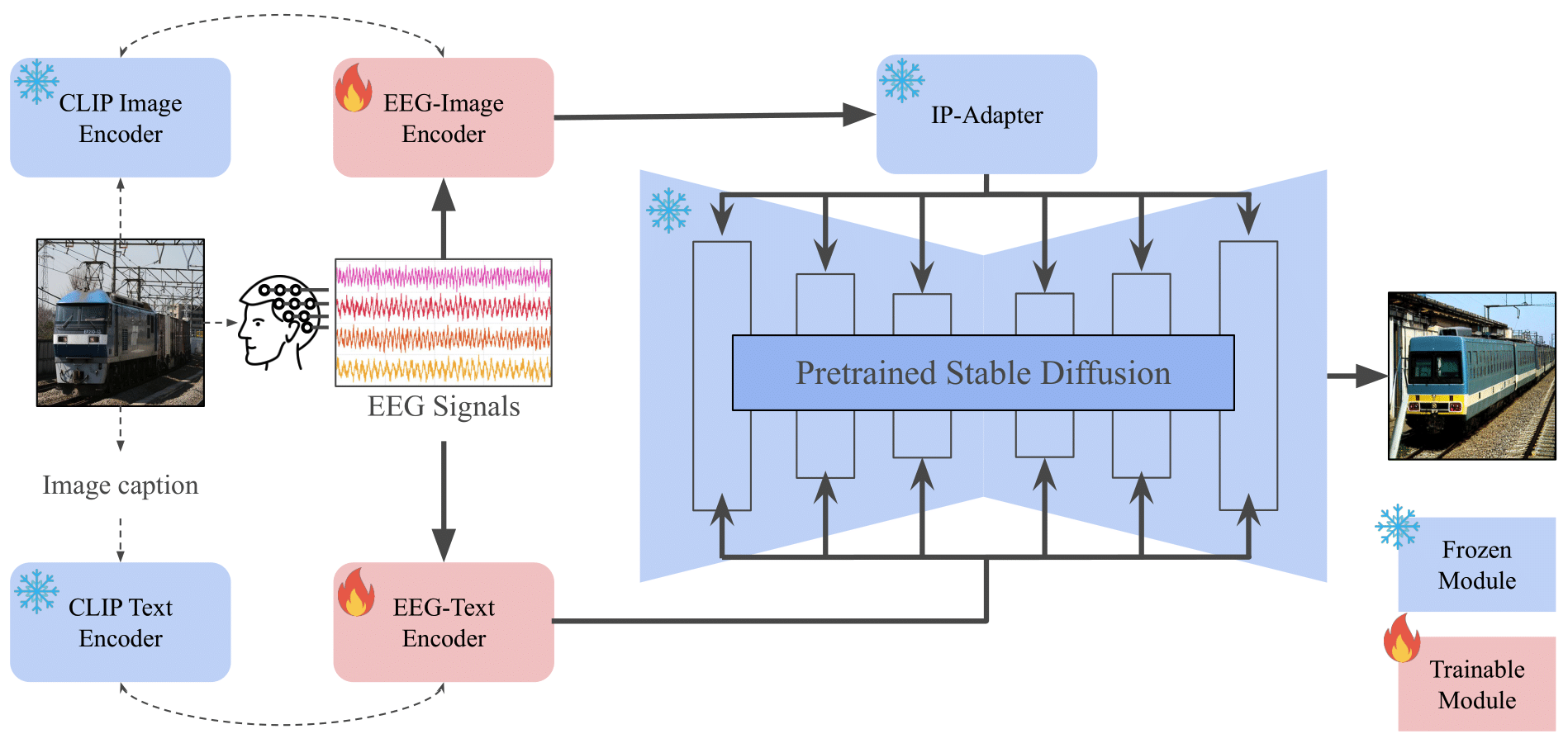}} % for arxiv
    \caption{The overall architecture of our proposed BrainDecoder framework. The modules in blue are frozen during training and only the modules in red are updated. The bold arrows are used during inference and the dotted lines are used during training.}
    \label{fig_framework}
\end{figure*}

\begin{figure*}[ht]
    % \centerline{\includegraphics[width=\textwidth]{Fig_SampleOutput.png}} % for icassp
    \centerline{\includegraphics[width=0.95\textwidth]{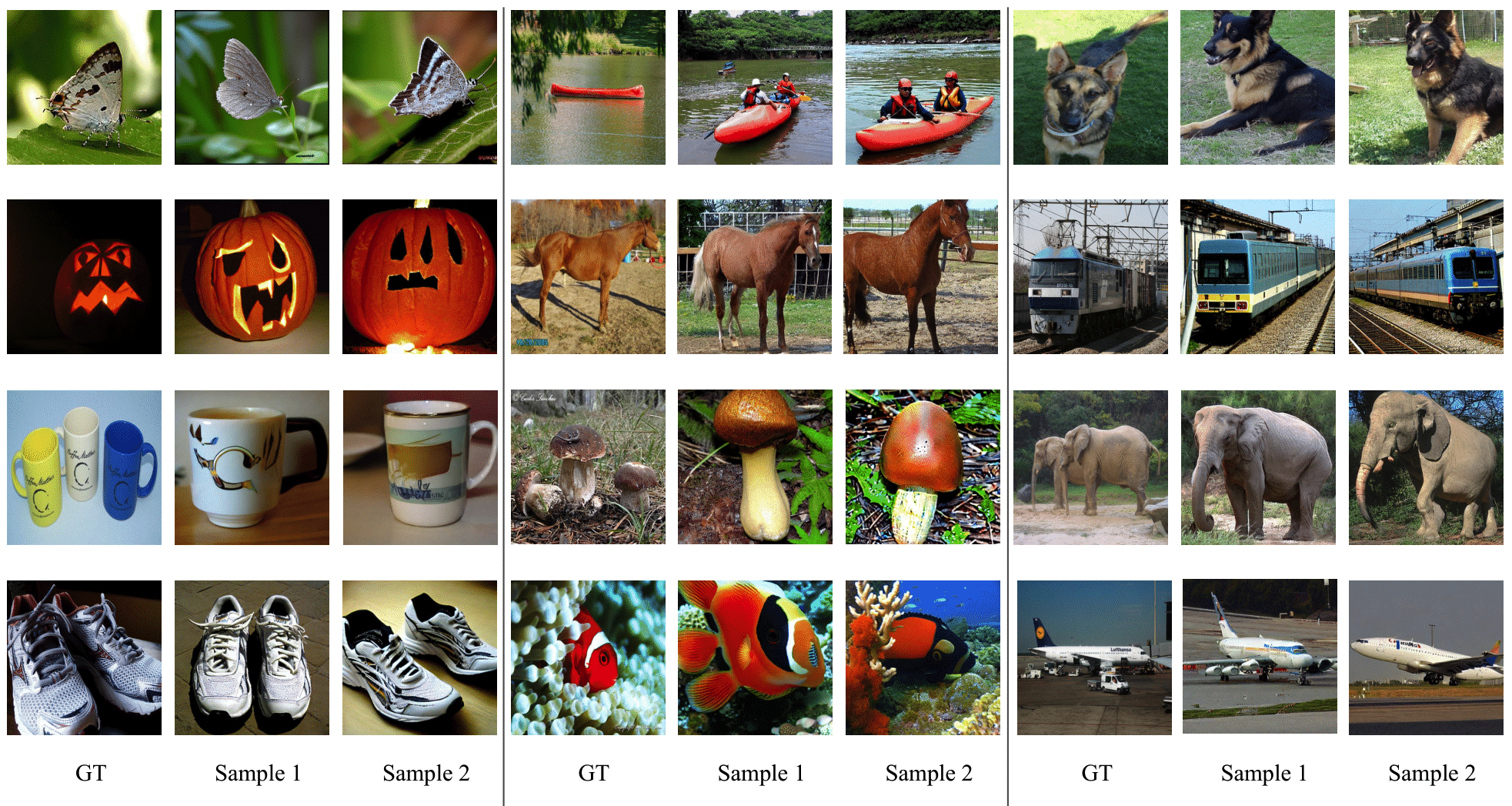}} % for arxiv
    \caption{Sample outputs. The images on the left show the ground truth visual stimuli shown during dataset collection. The following two images are sample outputs from our framework. Notably, the sample results show a high correspondence in semantics and style to the visual stimuli.}
    \label{fig_example_outputs}
\end{figure*}

\section{Methodology}
We introduce a novel framework for reconstructing images viewed by an EEG subject, as illustrated in Fig.~\ref{fig_framework}. It consists of three main components:
A) Aligning EEG signals with CLIP image space,
B) Aligning EEG signals with CLIP text space,
and C) Combining the CLIP-aligned EEG representations for visual stimuli reconstruction.

\subsection{EEG Alignment in Image Space}
Prior work \cite{ipadapter} \cite{imagedream} \cite{dalle2} has demonstrated the ability of CLIP image embeddings to facilitate both semantic and stylistic transfer when the generator model is conditioned accordingly. Building on these findings, our approach aims to extract image-related information from EEG signals by aligning them with CLIP image embeddings. To achieve this, we process the EEG signals and their corresponding ground truth images (i.e., the ones that the EEG subject was watching when the signal was taken) using an EEG encoder and a CLIP image encoder, respectively, and aim to correlate the outputs.
Let \(F_{\text{image}}\) be an encoder that processes the input EEG signal \(x\), and \(E_{\text{image}}\) the CLIP image encoder applied to the input image \(I\). We call \(F_{\text{image}}\) the EEG-image encoder because it is trained to align with the image as encoded by \(E_{\text{image}}\). We employ Mean Squared Error (MSE) as the loss function to measure the similarity between the EEG and image representations:
\begin{equation}
    L_{\text{image}} = \text{MSE}(F_\text{image}(x) - E_\text{image}(I)).
\end{equation}
% \begin{equation}
%     L_{\text{image}} = \text{MSE}(F_\mathrm{i}(x) - E_\mathrm{i}(I)).
% \end{equation}
To effectively encode the EEG data, we extend upon previous approaches \cite{brain2image} \cite{spampinato2017deep} \cite{eeg2image} by utilizing an LSTM-based encoder architecture followed by fully connected layers.

\subsection{EEG Alignment with Text Space}
Recent approaches for visual brain signal decoding  \cite{brainvis} \cite{braincaptioning} have sought to align brain signals with CLIP \cite{clip} text embeddings obtained from captions generated by pretrained image caption generators. However, since CLIP was trained on image-text pairs publicly available on the Internet with often short captions, those methods using longer generated captions, particularly with Stable Diffusion \cite{stablediffusion}, have been less effective. Although Stable Diffusion allows up to 77 tokens as input, empirical evidence suggests that the effective token length of CLIP is considerably shorter \cite{longclip}. Accordingly, we adopt a simpler labeling approach: we make the caption by appending the class label of the image to the text ``an image of''. We show empirically that this method improves performance over previous approaches and that more fine-grained information can be captured by the EEG-image encoder instead.
We use the CLIP text encoder \(E_{\text{text}}\) that embeds the caption $C$ to train the EEG-text encoder \(F_{\text{text}}\) that encodes the corresponding EEG signal \(x\). As in the image alignment step, we use MSE as the loss function to quantify the similarity between the EEG and the text representations:
\begin{equation}
    L_{\text{text}} = \text{MSE}(F_{\text{text}}(x) - E_{\text{text}}(C)).
\end{equation}
Similar to the image processing pipeline, an LSTM-based encoder is used for EEG signal encoding.

\subsection{Visual Stimuli Reconstruction}
After training the EEG-image and EEG-text encoders, we leverage the resulting EEG embeddings to generate images. Our method uses a pretrained latent diffusion model (e.g., Stable Diffusion \cite{stablediffusion}), with the EEG embeddings from both encoders serving as distinct conditioning inputs. This is achieved through a decoupled cross-attention mechanism \cite{ipadapter}. We hypothesize that by aligning the EEG signals in CLIP image space, the EEG encoder can capture detailed semantics and style that may not be easily conveyed through text alone. This approach is analogous to the way latent diffusion models incorporate both text and image prompts as conditioning factors.
The reconstructed visual stimuli are defined as:
\begin{equation}
    \hat{y} = \mathrm{SD}(F_i(x), F_t(x))
\end{equation}

Here, $\hat{y}$ represents the reconstructed image, and $\mathrm{SD}$ denotes the pretrained Stable Diffusion, conditioned on the outputs of both the EEG-image encoder \(F_{\text{image}}\) and the EEG-text encoder \(F_\text{text}\).

\section{Experiments and Results}
This section is divided into two main parts. We begin by detailing our experimental setup for training the EEG encoders. Following this, we present our findings and discuss various ablation studies.

\subsection{Dataset}
We utilize the Brain2Image \cite{spampinato2017deep} \cite{brain2image}, an EEG-image pair dataset with 11,466 EEG recordings from six participants, for our experiments. These recordings were captured using a 128-channel EEG system as the participants were exposed to visual stimuli for 500 ms. The stimuli consisted of 2,000 images with labels spanning 40 categories, derived from the ImageNet dataset \cite{imagenet}. Each category included 50 easily recognizable images to ensure clarity in the participants' neural responses.

%During data collection, each image was displayed for 0.5 seconds, and a pause was provided between sets of 50 images to prevent cognitive fatigue. The first 40 milliseconds of each EEG recording were discarded to remove any artifacts or noise associated with the initial response. We use the subsequent 440 milliseconds of data for analysis.

\subsection{Implementation}
For the EEG encoders, we extend from previous approaches \cite{spampinato2017deep} \cite{brain2image} \cite{eeg2image} and use a 3-layered LSTM network with a hidden dimension of 512. The output of the network is then passed through a fully connected linear network with a BatchNorm \cite{batchnorm} and LeakyReLU \cite{leakyrelu} activation function in between. Only the EEG encoders are trained in our framework, keeping the framework computationally efficient. We use the Adam \cite{adam} optimizer with a weight decay of 0.0001. The initial learning rate is set to 0.0003 and a lambda learning rate scheduler is used with a lambda factor of 0.999.

In order to align with CLIP image space, we follow the approach outlined in the IP-Adapter \cite{ipadapter} framework, utilizing the CLIP-Huge model to process the images. For aligning EEG with CLIP text space, we process the captions using the CLIP-Large model which is used by Stable Diffusion 1.5. The captions are generated by concatenating ``an image of'' with the class label. For ablation studies, we employ the LLaVA-1.5-7b model \cite{llava} for layout-oriented caption generation and BLIP \cite{blip} for general caption generation. 
For visual reconstruction, we employ Stable Diffusion version 1.5, aligning our method with recent results for fair comparison and we employ a PNDM scheduler \cite{pndm} with 25 inference steps.

\begin{figure}[t]
    \centerline{\includegraphics[width=\columnwidth]{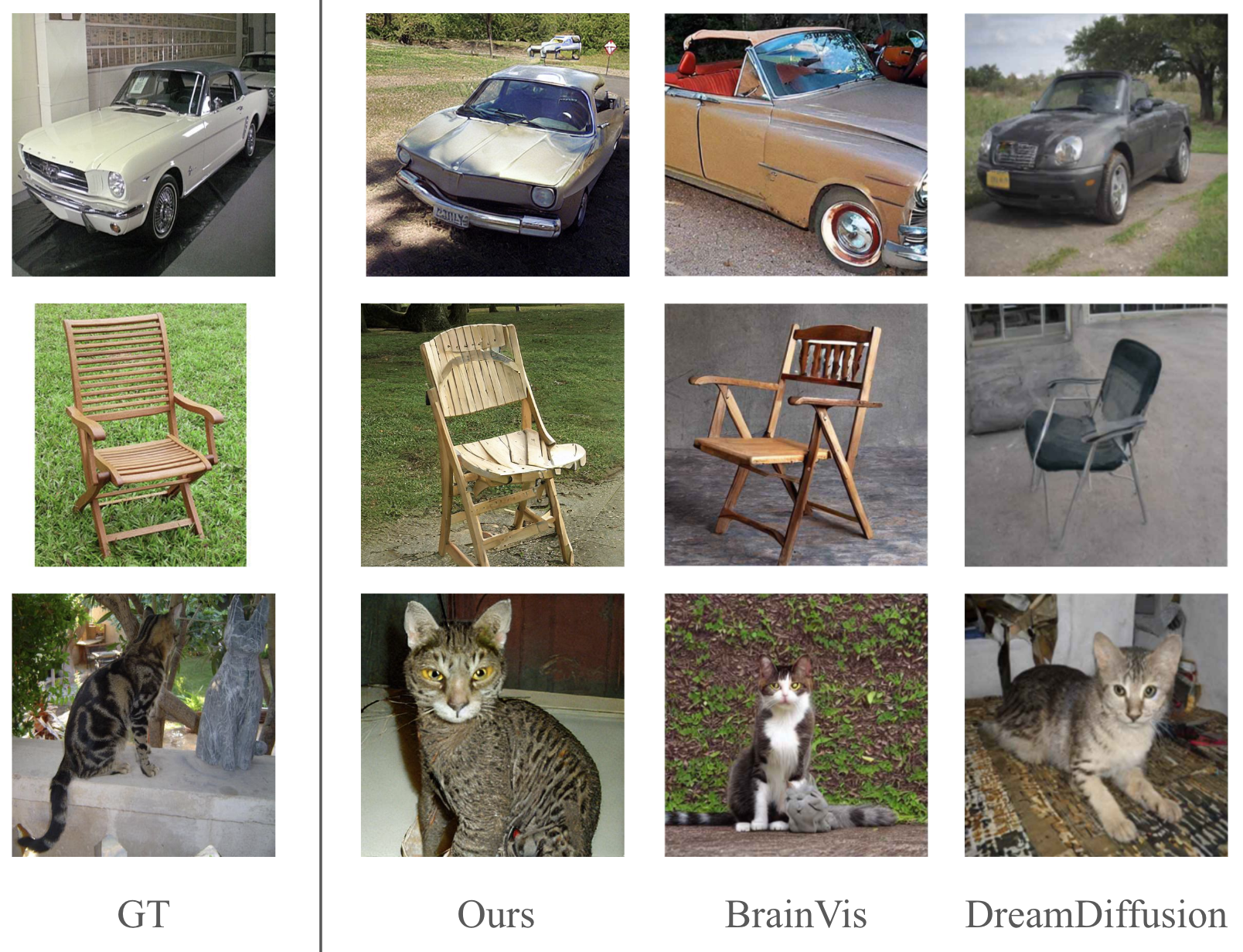}}
    \caption{Comparison of output images with the ground truth and outputs from other methods.}
    \label{fig_baseline_comparison}
\end{figure}

\subsection{Evaluation Metrics}\label{subsec:evalmetrics}
We employ the following metrics to objectively assess the performance of our framework.
{\bf ACC}: The $N$-way Top-$K$ Classification Accuracy \cite{mindvis} \cite{nwaytopkacc} evaluates the semantic accuracy of the reconstructed images. We set $N=50$ and $K=1$.
{\bf IS}: The Inception Score \cite{is} assesses the diversity and quality of the generated images.
{\bf FID}: Fréchet inception distance \cite{fid} measures the distance from the ground truth images.
{\bf SSIM}: The Structural Similarity Index Measure \cite{ssim} evaluates the quality of images.
{\bf CS}:  CLIP Similarity \cite{minddiffuser} \cite{brainvis} reflects how well the generated images capture the semantic and stylistic content of the ground truth images.

\subsection{Results}
% Qualitative
Fig.~\ref{fig_example_outputs} presents sample outputs of BrainDecoder. Beyond capturing the high-level semantics, our method demonstrates the ability to retain fine-grained visual features, including color and texture. Notably, there is also a resemblance in the color composition of the background in addition to the main object’s color. This capability is further illustrated in the example of the electric locomotive class. The object’s color is depicted as light blue---matching the visual stimuli---despite the range of potential color variations. This demonstrates the model's ability to recover nuanced visual attributes with a high fidelity.

This is further demonstrated in Fig.~\ref{fig_baseline_comparison}, where we compare our results with prior studies. Notably, in the second image, our method is able to reconstruct not only the wooden texture of the chair, but the grass in the background as well, which was absent in the results by other methods.

\begin{table}[t]
    \caption{Quantitative Results}
    \begin{center}
        \begin{tabular}{|c|c|c|c|c|c|}
            \hline
            \textbf{Methods} & \textbf{\textit{ACC $\uparrow$}} & \textbf{\textit{IS $\uparrow$}} & \textbf{\textit{FID $\downarrow$}} & \textbf{\textit{SSIM $\uparrow$}} & \textbf{\textit{CS $\uparrow$}}\\
            \hline
            Brain2Image & - & 5.07 & - & - & - \\
            DreamDiffusion$^{\mathrm{*}}$ & 45.8 & - & - & - & - \\
            BrainVis & 45.5 & - & - & - & 0.602 \\
            EEGStyleGAN-ADA & - & 10.82 & 174.13 & - & - \\
            \textbf{Ours} & \textbf{95.2} & \textbf{28.11} & \textbf{69.97} & \textbf{0.2277} & \textbf{0.7575} \\
            \hline
        \end{tabular}
        \label{table_quantitative_results}
    \end{center}
    \footnotesize{$^{\mathrm{*}}$Results from DreamDiffusion were computed using data from subject 4}
\end{table}

% Quantitative
Table. \ref{table_quantitative_results} shows the quantitative results of BrainDecoder compared to baselines \cite{brain2image} \cite{dreamdiffusion} \cite{brainvis} \cite{EEGStyleGAN-ADA}. We evaluate our methodology on 5 evaluation metrics in \textsection\ref{subsec:evalmetrics}. BrainDecoder outperforms the state-of-the-art in both reconstruction fidelity and generation quality. Notably, BrainDecoder achieves a surprising 95.2\% on the 50-way top-1 classification accuracy metrics, showing that the trained EEG encoders are able to extract rich information from the brain signals very well.

\begin{table}[t]
    \caption{Ablation Study Results}
    \begin{center}
        \begin{tabular}{|c|c|c|c|c|c|}
            \hline
            \textbf{Methods} & \textbf{\textit{ACC $\uparrow$}} & \textbf{\textit{IS $\uparrow$}} & \textbf{\textit{FID $\downarrow$}} & \textbf{\textit{SSIM $\uparrow$}} & \textbf{\textit{CS $\uparrow$}}\\
            \hline
            Only Text (LLaVA) & 60.43 & 20.22 & 151.98 & 0.1797 & 0.6188 \\
            Only Text (BLIP) & 68.91 & 24.0 & 127.19 & 0.1832 & 0.6541 \\
            Only Text (label) & 72.61 & 26.43 & 105.6 & 0.1845 & 0.6610 \\
            Only Image & 79.7 & 26.1 & 75.88 & 0.2239 & 0.7177 \\
            \textbf{Original} & \textbf{95.2} & \textbf{28.11} & \textbf{69.97} & \textbf{0.2277} & \textbf{0.7575} \\
            \hline
        \end{tabular}
        \label{table_ablation_study}
    \end{center}
\end{table}

\subsection{Ablation}
We conduct an ablation study to understand the contributions of each component. Rows 3-5 of Table~\ref{table_ablation_study} show visual decoding with the EEG-image encoder yields a higher SSIM (0.2239) than with only the EEG-text encoder (0.1845). This supports our premise that aligning in CLIP’s image space facilitates style transfer. Furthermore, the framework achieves the best performance when both encoders are used, indicating the complementary nature of the two encoders.

\begin{figure}[t]
    \centerline{\includegraphics[width=\columnwidth]{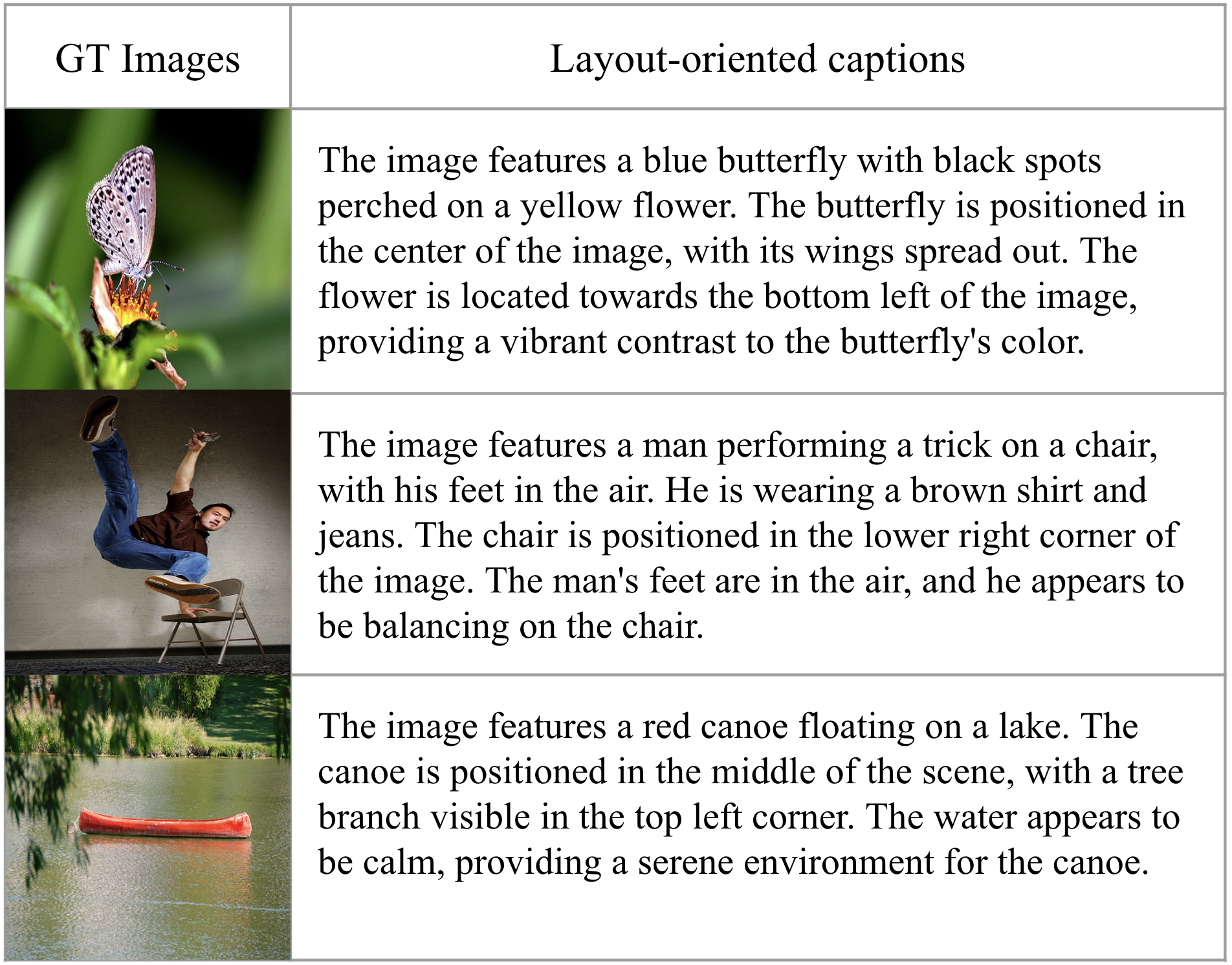}}
    \caption{Example layout-oriented captions generated with LLaVA.}
    \label{fig_example_layout_caption}
\end{figure}

Additionally, we empirically show that captions generated by Vision Language Models (VLMs) are suboptimal for EEG-based visual decoding. We compare our label caption method with two VLMs: BLIP \cite{blip} and layout-oriented LLaVA \cite{llava}. A key challenge in image reconstruction from brain signals is preserving the visual layout. We hypothesize that associating EEG signals with detailed layout-oriented CLIP text embeddings might help. Using the LLaVA model, we generate layout-oriented captions following the instruction: ``Write a description of the image layout. EXAMPLE OUTPUT: [object] is in the top left of the image, facing right.'' Fig.~\ref{fig_example_layout_caption} shows example layout-oriented captions. Notably, rows 1-3 of Table \ref{table_ablation_study} indicate that the simple label caption (``an image of [class name]'') performs best, while layout-oriented captions (row 1) perform the worst. This further supports our premise that simple label captions are more effective for EEG encoders and complex prompts are harder for CLIP to fully interpret.

% \subsection{Limitations}
% figure
% count
% text limited information
% direction
% small dataset
% individual variance

\section{Conclusion}

% Our research introduces \textit{\textbf{BrainDecoder}}, a novel approach to image reconstruction from EEG signals, with a key focus on preserving stylistic and perceptual features of the original visual stimuli. BrainDecoder integrates a unique framework that separately aligns EEG signals with both CLIP image embeddings and CLIP text embeddings. By leveraging this dual-alignment process, we bridge the gap between neural representations and both visual content and its higher-level semantic associations. 
% Our quantitative and qualitative analysis highlight significant improvements over existing models revealing that this dual-alignment strategy offers a richer and more comprehensive interpretation of neural signals.

Our research introduces \textit{\textbf{BrainDecoder}}, a novel approach to image reconstruction from EEG signals that preserves both stylistic and semantic features of visual stimuli. By aligning EEG signals with CLIP image and text embeddings separately, we bridge the gap between neural representations and visual content. Our analysis demonstrates significant improvements over existing models, offering a richer interpretation of neural signals through the dual-alignment strategy.
% resulting in reconstructed images that better capture the nuances of the original visual stimuli.

\clearpage
\bibliography{references}
% \section*{References}

% Unless there are six authors or more give all authors' names; do not use
% ``et al.''. Papers that have not been published, even if they have been
% submitted for publication, should be cited as ``unpublished'' \cite{b4}. Papers
% that have been accepted for publication should be cited as ``in press'' \cite{b5}.
% Capitalize only the first word in a paper title, except for proper nouns and
% element symbols.

% \vspace{12pt}
% \color{red}
% IEEE conference templates contain guidance text for composing and formatting conference papers. Please ensure that all template text is removed from your conference paper prior to submission to the conference. Failure to remove the template text from your paper may result in your paper not being published.

\end{document}